\documentclass[conference]{IEEEtran}
\IEEEoverridecommandlockouts
\usepackage{cite}
\usepackage{amsmath,amssymb,amsfonts}
\usepackage{graphicx}
\usepackage{textcomp}
\usepackage{booktabs}
\usepackage{xcolor}
\usepackage{mathtools}
\usepackage[nolist]{acronym}
\usepackage{comment}
\usepackage{bbold}
\usepackage{enumitem}
\usepackage{units}
\usepackage{tabularx}
\usepackage{microtype}
\usepackage{algorithm}
\usepackage[noend]{algpseudocode}
\usepackage{etoolbox}

\algrenewcommand\algorithmicindent{1em}%

\DeclareMathOperator{\svd}{SVD}
\DeclareMathOperator{\Euler}{Euler}
\newcommand\Algphase[1]{%
\vspace*{-.7\baselineskip}\Statex\hspace*{\dimexpr-\algorithmicindent-2pt\relax}\rule{0.97\columnwidth}{0.2pt}%
\Statex\hspace*{-\algorithmicindent}\textbf{#1}%
\vspace*{-.7\baselineskip}\Statex\hspace*{\dimexpr-\algorithmicindent-2pt\relax}\rule{0.97\columnwidth}{0.2pt}%
}

\definecolor{todo-red}{RGB}{200,12,12}
\definecolor{green4}{RGB}{0,128,0}

\def\BibTeX{{\rm B\kern-.05em{\sc i\kern-.025em b}\kern-.08em
    T\kern-.1667em\lower.7ex\hbox{E}\kern-.125emX}}

\newcommand{\etal}{\textit{et al}. }
\newcommand{\arrow}{$\rightarrow\;$}

\DeclareMathOperator*{\argmin}{arg\,min}

\begin{document}

\title{Superquadric Object Representation for Optimization-based Semantic SLAM
\thanks{This work was partially supported by Siemens Mobility, Germany, and the ETH Mobility Initiative under the projects \textit{PROMPT} and \textit{LROD}.}
}

\author{\IEEEauthorblockN{Florian Tschopp}
\IEEEauthorblockA{\textit{Autonomous Systems Lab} \\
\textit{ETH Zurich}\\
Zurich, Switzerland \\
ftschopp@ethz.ch}
\and
\IEEEauthorblockN{Juan Nieto}
\IEEEauthorblockA{\textit{Mixed Reality and AI} \\
\textit{Microsoft Switzerland}\\
Zurich, Switzerland \\
juannieto@microsoft.com}
\and
\IEEEauthorblockN{Roland Siegwart}
\IEEEauthorblockA{\textit{Autonomous Systems Lab} \\
\textit{ETH Zurich}\\
Zurich, Switzerland \\
rsiegwart@ethz.ch}
\and
\IEEEauthorblockN{Cesar Cadena}
\IEEEauthorblockA{\textit{Autonomous Systems Lab} \\
\textit{ETH Zurich}\\
Zurich, Switzerland \\
cesarc@ethz.ch}
}

\maketitle

\begin{abstract}
Introducing semantically meaningful objects to visual \ac{slam} has the potential to improve both the accuracy and reliability of pose estimates, especially in challenging scenarios with significant viewpoint and appearance changes. 
However, how semantic objects should be represented for an efficient inclusion in optimization-based \ac{slam} frameworks is still an open question.
\Acp{sq} are an efficient and compact object representation, able to represent most common object types to a high degree, and typically retrieved from 3D point-cloud data. 
However, accurate 3D point-cloud data might not be available in all applications.
Recent advancements in machine learning enabled robust object recognition and semantic mask measurements from camera images under many different appearance conditions.
We propose a pipeline to leverage such semantic mask measurements to fit \ac{sq} parameters to multi-view camera observations using a multi-stage initialization and optimization procedure.
We demonstrate the system's ability to retrieve randomly generated \ac{sq} parameters from multi-view mask observations in preliminary simulation experiments and evaluate different initialization stages and cost functions.
\end{abstract}
\acresetall
\section{Introduction}
Determining a system's position in its environment is a crucial task for most mobile robotic applications.
Examples include navigation for both wheeled~\cite{Gawel2019AConstruction} and flying~\cite{Blosch2010VisionEnvironments} robots, autonomous driving~\cite{Burki2019}, and \ac{adas} for cars and trains~\cite{Tschopp2019ExperimentalVehicles}.
Such positioning is often tackled using visual \ac{slam} techniques~\cite{RolandSiegwart2011}, achieving robust and accurate state estimation in many applications~\cite{Schneider2017,Mur-Artal2015}.
However, as most of these systems rely on appearance-based landmarks such as BRISK~\cite{Leutenegger2011} or ORB~\cite{Rublee}, their accuracy and reliability tend to decrease in case of significant appearance change of the environment due to variations in illumination or viewpoint. 
Such changes are especially evident in outdoor applications, in which daytime, weather, and seasonal conditions can influence the appearance significantly~\cite{Milford2012, Sattler2018BenchmarkingConditions}.
In addition, typical sparse~\cite{Schneider2017, Mur-Artal2017} or semi-dense~\cite{Engel2018DirectOdometry} mapping approaches rely on many \textit{weak}\footnote{Weak landmarks feature limited distinctness in contrast to \textit{strong} landmarks, which are very distinct. However, in combination with many other weak landmarks and geometric verification, weak landmarks can achieve good results.} landmarks to enable crucial functionalities such as loop-closure and 6~\ac{dof} localization.
Therefore, the number of landmarks required for large-scale mapping might invalidate some of the approaches due to memory and bandwidth constraints~\cite{Burki2019}.
\begin{figure}
    \centering
    \includegraphics[width=\columnwidth]{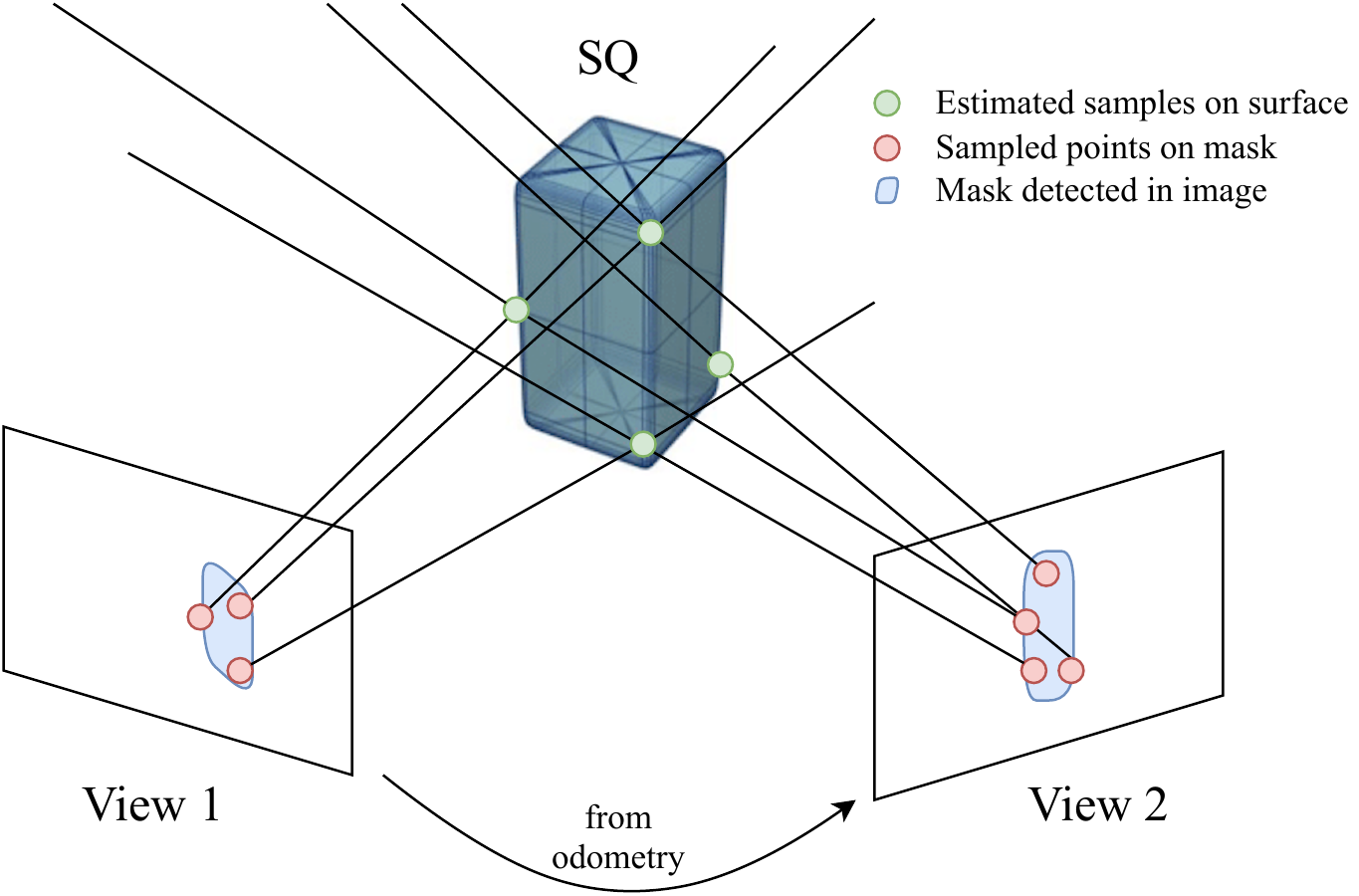}
    \caption{Superquadric parameter fitting from multi-view semantic mask object observations using non-linear optimization.}
    \label{fig:overview}
\end{figure}

Over the last decade, \ac{dl}-based semantic segmentation and object detection algorithms~\cite{He2017,Bolya2020YOLACT++:Segmentation,Redmon2016YouDetection} have steadily improved and achieved robust object recognition and instance segmentation covering many different appearance conditions. 
To leverage these advancements, instead of including generic geometric primitives as landmarks, such as keypoints~\cite{Schneider2017, Mur-Artal2017} or lines~\cite{Verhagen2014Scale-invariantMatching,Lee2014OutdoorLines,Micusik2015DescriptorSegments}, or utilizing global image descriptors~\cite{Oliva2006BuildingGlob,Arandjelovic2013AllVLAD,Arandjelovic2018NetVLAD:Recognition} for finding previously visited places, including semantic understanding for \ac{slam} and localization can significantly improve the performance~\cite{Gawel2017,Taubner2020LCDRecognition,Schonberger2017,Cramariuc2021SemSegMapLocalization}.
One way to include semantic understanding is by mapping semantic objects as proposed by Nicholson~\etal~\cite{Nicholson2019QuadricSLAM:SLAM} or Frey~\etal~\cite{Frey2019EfficientSLAM} as \textit{strong} landmarks. 
Semantic objects can be consistently detected, are frequent permitting that localization can be achieved often, but are also compact and sparse, enabling large-scale mapping.
Furthermore, semantic objects have the potential to be highly descriptive to facilitate localization independent of the current appearance leading to robustness against viewpoint, seasonal, weather, or daytime changes~\cite{Sattler2018BenchmarkingConditions}.

However, in order to utilize semantic landmarks in state-of-the-art optimization-based \ac{slam} systems~\cite{Strasdat2010}, on top of finding a distinctive and robust descriptor~\cite{Taubner2020LCDRecognition}, a compact and computationally efficient representation of the objects has to be found, providing a high representation strength, i.e. is able to accurately represent most common object types.

In this paper, we propose to use \acp{sq} as semantic object representations, as a compromise between compactness and representation strength.
To enable the usage of \acp{sq} in mapping tasks without being dependent on depth data, we propose to retrieve \ac{sq} parameters from multi-view semantic mask observations as shown in Figure~\ref{fig:overview}.
Furthermore, we propose the adaption of an analytic cost function of the fitting quality to multi-view mask observations for an efficient implementation of \acp{sq} in optimization-based \ac{slam} frameworks.
We evaluate the retrieval quality and multiple initialization techniques on randomly generated \acp{sq} in simulation, demonstrating the ability to successfully retrieve \ac{sq} parameters with high fitting accuracy.

\section{Related Work}
Semantic objects can be mathematically represented in many different ways, typically as a trade-off between parametrization complexity and representation strength.
Simple 3D point representations as mapped in~\cite{Tschopp2021Hough2Map, Frey2019EfficientSLAM} consist of three position parameters, equivalent to standard keypoints.
This benefits from an easy integration of well-known re-projection cost functions, but especially objects with larger sizes or non-spherical shapes cannot be represented accurately.
This problem can be addressed by including size parameters resulting in cubes~\cite{Yang2019CubeSLAM:SLAM}, spheres~\cite{Papadakis2018}, or ellipsoids/quadrics~\cite{Ok2019RobustNavigation,Nicholson2019QuadricSLAM:SLAM}. 
However, if the mapped object has a different shape than modeled by the representation, inaccuracies and wrong scene understandings might result.
In contrast, complex dense representations such as \acp{esdf}~\cite{Mccormac2018Fusion++:Map,Grinvald2019,Strecke2019EM-Fusion:Association,Rosinol2020Kimera:Mapping} result in measurement functions that are tricky to include in optimization frameworks, rely on a high dimensional parametrization, and require many observations for a good shape estimation.
Ultimately, the object can be represented by high-accuracy 3D models~\cite{Galvez-Lopez2016Real-timeSLAM,Salas-Moreno2013SLAM++:Objects}, which, however, requires a good database of the expected objects in order to work reliably.

\acp{sq} as initially introduced to the computer vision community by Barr~\etal~\cite{Barr1981SuperquadricsTransformations} are an extension to standard quadrics and can represent a wide range of common convex object types with only 11 parameters.
The retrieval of \ac{sq} parameters is extensively researched and typically solved by fitting 3D point-cloud data using non-linear least-squares optimization~\cite{Boult1988,Wu1994,Zhang2003,Duncan2013,Makhal2017,Vaskevicius2019}.
Also, \ac{dl}-based retrieval from 3D point-cloud data was recently proposed~\cite{Paschalidou2019}.
In contrast, in this work, we focus on retrieving \ac{sq} parameters from multi-view camera observations to be independent on accurate range data, which might not be available, especially in outdoor environments.

Most object-based \ac{slam} frameworks, which do not depend on depth data, retrieve and optimize object and camera pose parameters from 2D bounding box observations~\cite{Nicholson2019QuadricSLAM:SLAM,Ok2019RobustNavigation,Yang2019CubeSLAM:SLAM}.
However, such camera-frame axis-aligned bounding boxes are not able to accurately represent objects which are not aligned to the camera view, especially if the objects have non-equal dimensions.
In contrast, recent semantic instance segmentation networks~\cite{He2017,Bolya2020YOLACT++:Segmentation} achieve high accuracy in not only detecting the objects but also capturing their shape in the current camera view.

How such semantic mask observations can be utilized to efficiently retrieve and optimize \ac{sq} parameters and leverage the additional shape information is an open question that we aim to address in this paper.

\section{Superquadric Fitting for Semantic Measurements} \label{sec:method}
This section gives an overview of \acp{sq}, introduces our pipeline to retrieve and optimize \ac{sq} parameters based on semantic mask observations, and describes an analytic cost function that approximates the fit of the observation data with the \ac{sq} model.
\subsection{Notation}
In this document, we denote scalars as $a$, vector-valued variables as $\mathbf{a}$, matrices as $\mathrm{A}$, and a set of variables as $\mathcal{A}$.
3D points are denoted in homogeneous coordinates as $\prescript{}{A}{\mathbf{t}} \in \mathbb{R}^4$ represented in the coordinate frame $A$, and a set of 3D points as $\prescript{}{A}{\mathcal{T}}$.
Furthermore, $\mathrm{T}^B_A \in \mathbb{R}^{4\times4}$ indicates a homogeneous coordinate transformation to transform a point described in frame $B$ into frame $A$, i.e. $\prescript{}{A}{\mathbf{t}} = \mathrm{T}^B_A \cdot \prescript{}{B}{\mathbf{t}}$.
Poses of both cameras and \acp{sq} are denoted by a position $\prescript{}{W}{\mathbf{p}} \in \mathbb{R}^3$ and an orientation $\prescript{}{W}{\mathbf{r}} \in \mathbb{R}^3$ using Euler angles with respect to the world coordinate frame $W$.
A set of poses is denoted by $\mathcal{P}$.

\subsection{Superquadrics}

\Acp{sq} are mathematical shapes fully describable by a compact number of parameters.
They extend standard quadrics by incorporating additional shape parameters to define the objects' roundness.
A point on the surface of a \ac{sq} $\prescript{}{SQ}{\mathbf{t}} = \left[t_x, t_y, t_z\right]$ is found by its direct formulation~\cite{Boult1988}
\begin{equation} \label{eq:explicit}
    \prescript{}{SQ}{\mathbf{t}} = \left[ \begin{array}{c} a_x \cdot \cos(\eta)^{\varepsilon_1}\cdot \cos(\omega)^{\varepsilon_2} \\
    a_y \cdot \cos(\eta)^{\varepsilon_1}\cdot \sin(\omega)^{\varepsilon_2}\\
    a_z \cdot \sin(\eta)^{\varepsilon_1} 
    \end{array} \right], 
\end{equation}
where $a_{\bullet}$ are the size parameters in each dimension, $\varepsilon_\bullet$ are the two shape parameters, and $-\frac{\pi}{2}\leq\eta\leq\frac{\pi}{2}$ and $-\pi\leq\omega\leq\pi$ are iteration variables. 
Using Equation~(\ref{eq:explicit}), points on the \ac{sq} surface can be sampled and projected to the camera view to evaluate the quality of fit between the projected \ac{sq} and the mask observations by evaluating the \ac{riou}.

However, in this work, primarily the implicit formulation of \acp{sq}~\cite{Jaklic2000} is of interest given by
\begin{equation} \label{eq:implicit}
    F\left(\prescript{}{SQ}{\mathbf{t}}\right) = \left(\left(\frac{t_x}{a_x}\right)^{\frac{2}{\varepsilon_2}} + \left(\frac{t_y}{a_y}\right)^{\frac{2}{\varepsilon_2}}\right)^{\frac{\varepsilon_2}{\varepsilon_1}} + \left(\frac{t_z}{a_z}\right)^{\frac{2}{\varepsilon_1}} = 1,
\end{equation}
where $\prescript{}{SQ}{\mathbf{t}} = \left[t_x, t_y, t_z\right]$ is a point on the \ac{sq} surface in \ac{sq} coordinates.

In addition to the size $\mathbf{a}$ and shape $\mathbf{\varepsilon}$, a general \ac{sq} in 3D space is defined by its position $\prescript{}{W}{\mathbf{p}_{SQ}}\in \mathbb{R}^3$ and orientation $\prescript{}{W}{\mathbf{r}_{SQ}}\in\mathbb{R}^3$ in world coordinates $W$, forming a transformation $\mathrm{T}_{SQ}^{W}$ from world to \ac{sq} coordinates. 
In total, a \ac{sq} is defined by the parameters $\mathbf{\xi} = \left[ \mathbf{a}, \mathbf{\varepsilon}, \mathbf{p}, \mathbf{r}\right]\in\mathbb{R}^{11}$.
The implicit formulation in Equation~(\ref{eq:implicit}) directly provides an insight into whether a point $\prescript{}{SQ}{\mathbf{t}}$ lies on the surface $F\left(\prescript{}{SQ}{\mathbf{t}}\right) = 1$, is located outside of the \ac{sq} $F\left(\prescript{}{SQ}{\mathbf{t}}\right) > 1$, or on its inside $F\left(\prescript{}{SQ}{\mathbf{t}}\right) < 1$.
\begin{figure}
    \centering
    \includegraphics[clip, trim=1cm 0cm 1.5cm 0cm, width=\columnwidth]{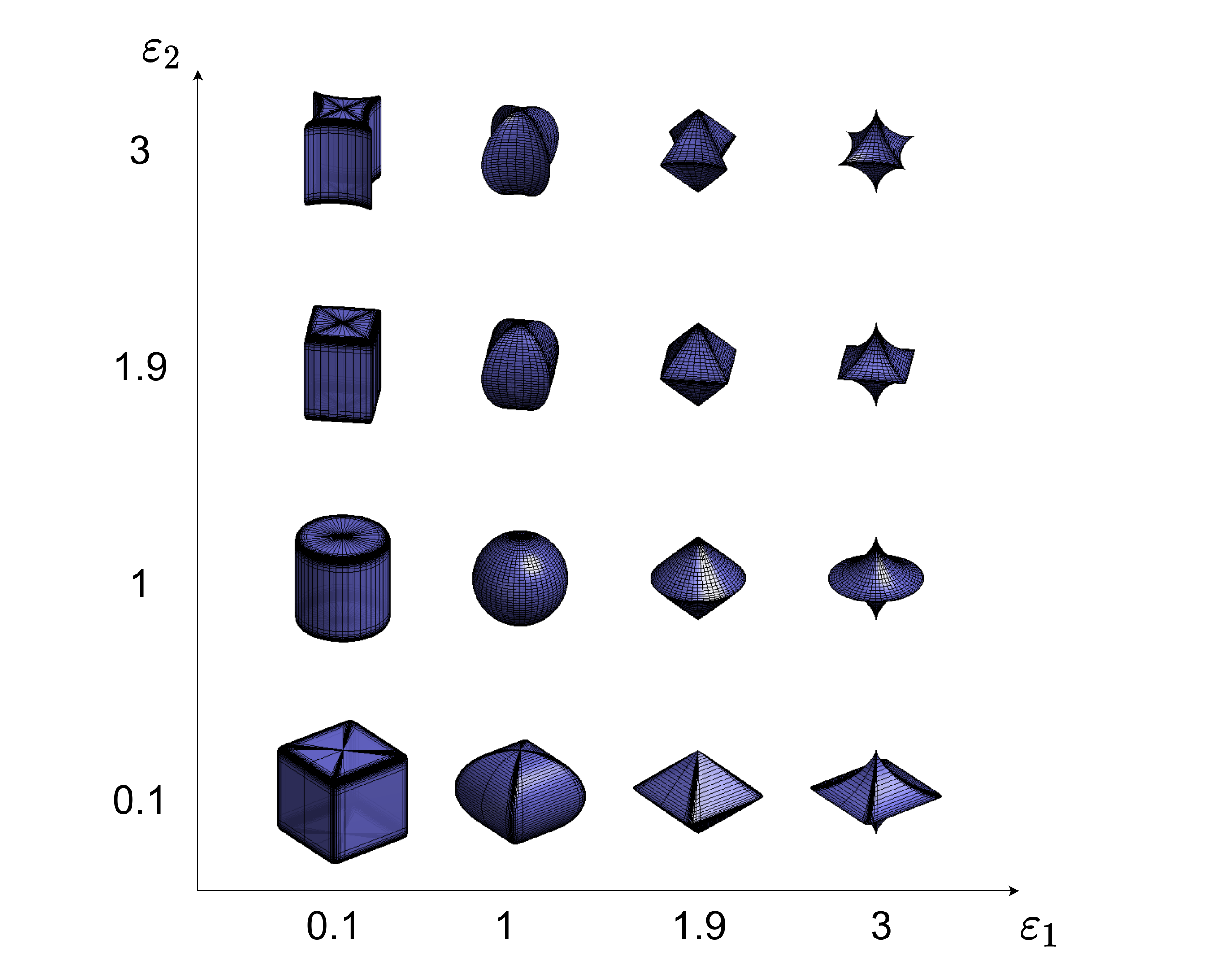}
    \caption{\acp{sq} with unit size ($\mathbf{a}=\mathbf{1}$). The shape parameters $\mathbf{\varepsilon}$ define the \textit{edginess} of the final object. To keep the optimization stable~\cite{Vaskevicius2019} and consider only convex shapes, only shape parameters $0.1 \leq \mathbf{\varepsilon} \leq 1.9$ are considered in this work.}
    \label{fig:shape_param}
\end{figure}%

Depending on the coupling of the shape parameters $\varepsilon$, \acp{sq} represent cubic objects, spheres, and ellipsoids, up to convex shapes and everything in between, such as cylinders.
Figure~\ref{fig:shape_param} shows an overview of the achievable shapes with different shape parameters.
In the case where both shape parameters are $\varepsilon = \mathbf{1}$, the \ac{sq} becomes equivalent to a standard quadric.
In this work, to minimize numerical problems and keep the optimization as stable as possible~\cite{Vaskevicius2019}, we only consider convex objects with shape parameters
\begin{equation} \label{eq:eps_const}
    0.1 \leq \mathbf{\varepsilon} \leq 1.9.
\end{equation}

\subsection{Optimizing \ac{riou}} \label{sec:riou}
To achieve our goal of \ac{sq} parameter retrieval given observation data, in our case multi-view semantic mask observations $\mathcal{O}$, we formulate the following optimization problem
\begin{equation} \label{eq:opt_riou}
     \xi_{opt} = \argmin_{\xi}{\sum_{p=1}^P}G_1\left(\xi,\mathcal{O}, \mathcal{P}\right)^2,
\end{equation}
where $\mathcal{P}$ is the set of $P$ poses that observe the \ac{sq}.
This optimization problem can be solved using the non-linear least-squares Levenberg-Marquardt optimization algorithm~\cite{Levenberg1944ASquares,Marquardt1963AnParameters}.
The cost function $G_1$ is formed by 
\begin{equation}
    G_1 = 1-\ac{riou}\left(\xi,\mathcal{O}, \mathcal{P}\right),
\end{equation}
where the \ac{riou} is evaluated by comparing the re-projected estimated \ac{sq} $\xi$ according to Equation~(\ref{eq:explicit}) and camera poses $\mathcal{P}$ with the semantic mask observation $o \in \mathcal{O}$.
As the mask observation is likely to be non-parametric, the 2D \ac{iou} evaluation is based on polyshapes~\cite{MathWorksSchweizPolyshapeMATLAB}.
This drastically complicates the retrieval of analytic Jacobians for the cost function.
Therefore, numerical derivations using finite differences are used for the optimization of $G_1$. 
The state constraints $\xi_{th}$ mentioned in Equation~(\ref{eq:eps_const}), as well as the condition of a non-negative and non-vanishing size $\mathbf{a} \geq 0.1$, are included in the optimization as a soft constraint penalty
\begin{equation} \label{eq:constraints}
\begin{aligned}
    G_2 &=  G_1 + c_p \cdot \sum_{i=1}^K \rho_i,\\
    \rho_i &= \begin{dcases}  |\xi^i - \xi_{th}^i| & \text{if} \; \xi^i < \xi_{th,l}^i \; \text{or}\; \xi^i > \xi_{th,u}^i\\
    0& \text{otherwise}, 
    \end{dcases}
\end{aligned}
\end{equation}
where $c_p$ is a heuristic cost penalty, $\xi^i$ is one of the state variables, $K$ is the state size, and $\xi_{th,l}$ and $\xi_{th,u}$ are the lower and upper thresholds, respectively.
\subsection{Analytic cost function and optimization problem}
\label{sec:cost_anaytic}
To effectively use \ac{sq}-landmarks in the optimization of a factor-graph-based \ac{slam} setup, an efficient and derivable formulation of the \ac{riou} is required.
The \textit{radial distance} $G_3$ was proposed by Zhang~\etal~\cite{Zhang2003} to represent the error of a \ac{sq} fitted to a 3D \ac{pc}.
$G_3$ is the distance from any given point $_W\mathbf{t}$ to the surface of a \ac{sq} $\xi$ and is calculated as
\begin{equation}
\label{sq:rad}
G_3\left(\mathbf{\xi}, \prescript{}{SQ}{\mathbf{t}}\right) = ||\prescript{}{SQ}{\mathbf{t}}|| \left[F(\prescript{}{SQ}{\mathbf{t}})^{-\frac{\varepsilon_1}{2}} - 1 \right],
\end{equation}
where $\prescript{}{SQ}{\mathbf{t}} = \mathrm{T}_{SQ}^{W}\cdot \prescript{}{W}{\mathbf{t}}$ is a 3D point transformed into \ac{sq} coordinates and $F$ is the implicit \ac{sq} formulation in Equation~(\ref{eq:implicit}).
The main objective for \ac{sq} fitting, as shown in e.g.~\cite{Wu1994, Vaskevicius2019, Zhang2003, Duncan2013, Makhal2017} is then to minimize $G_3$ for a number $N$ of 3D point observations such that
\begin{equation}
\label{eq:opt}
    \xi_{opt} = \argmin_{\xi}{\sum_{n=1}^N}G_3\left(\xi,\mathrm{T}_{SQ}^{W}\cdot\prescript{}{W}{\mathbf{t}_n}\right)^2.
\end{equation}
However, our approach aims to recover \ac{sq} shapes from multi-view camera mask observations without being dependent on accurate depth data.
Therefore, such mask observations are randomly sampled to obtain $N$ observation samples $\mathbf{s} = \left[s_x, s_y\right] \in \mathcal{S}_o$ per object observation $o \in \mathcal{O}$. 
Finally, $\mathbf{s}$ can be back-projected from the estimated camera pose $\mathrm{T}_{W}^{C}\left(\mathbf{p}_C, \mathbf{r}_C \right) \in \mathcal{P}$, where $\mathbf{p}_C$ and $\mathbf{r}_C$ are the camera position and orientation, respectively, 
\begin{equation}\label{eq:backprojection}
\begin{split}
    \prescript{}{W}{\mathbf{t}} &= {T}_{W}^{C} \cdot Bp(\mathbf{s}, d) \\
    &= \mathrm{T}_{W}^{C} \cdot \left[ d\cdot \left[(s_x - \kappa_x) / f_x, (s_y - \kappa_y) / f_y,  1\right]^\top \right],
\end{split}
\end{equation} 
using known camera intrinsics, where $\kappa_\bullet$ are the camera centers and $f_\bullet$ are the focal lengths.
The parameters $d \in \mathcal{D}$ are unknown depth parameters and can either be per camera view termed \textit{combined} depth, or per single observation sample $\mathbf{s}$ termed \textit{separate} depth.
Hence, the optimization problem of Equation~(\ref{eq:opt}) becomes
\begin{equation} \label{eq:opt_d}
    \left[ \xi_{opt}, \mathcal{D}_{opt} \right] = \argmin_{\xi, \mathcal{D}}{\sum_{p=1}^P\sum_{n=1}^N}G_4\left(\xi, \mathcal{S}_{p,n}, \mathcal{P}_p, \mathcal{D}_p\right)^2.
\end{equation}

$G_4$ cannot penalty \acp{sq} that are larger than what is represented by the mask observations, as long as the sample points $\mathcal{S}$ are on its surface.
Together with the variable depths $d \in \mathcal{D}$, this results in an unconstrained size of the \ac{sq} which is typically heavily overestimated.
To circumvent this issue, an additional factor is introduced to the cost function to retrieve the minimal size \ac{sq}, which still agrees to the mask observation data
\begin{equation}
    G_5 = \left(a_x\cdot a_y \cdot a_z + 1\right) \cdot G_4.
\end{equation}

For all analytic cost functions $G_3-G_5$, the same constraint violation penalty introduced in Equation~(\ref{eq:constraints}) is applied.

\subsection{Multi-stage optimization}
\label{sec:multi-stage}
Using the Levenberg-Marquardt algorithm to optimize $G_5$, we discovered that the optimization is fragile and subject to deep local minima.
Therefore, to achieve high robustness, reliability, and accuracy, a good initialization of all \ac{sq} parameters close to the global minimum is required.
We propose to use the following multi-stage initialization and optimization procedure to achieve good convergence of the parameters:
\subsubsection{Triangulation and combined depth}
Triangulation is a widespread method typically used to initialize 3D keypoint positions from multi-camera observations.
We use linear triangulation to initialize the position of the \ac{sq} $\prescript{}{W}{\mathbf{\hat{p}}}_{SQ}$ based on back-projection of the centroid $\mathbf{c}_i$ of the mask observations
\begin{equation}
    \prescript{}{W}{\mathbf{v}}_i =  \mathrm{R}^{C_i}_W \cdot Bp(\mathbf{c}_i, 1) \quad \forall \,i \in \mathcal{P}, 
\end{equation}
\begin{equation}
    \mathrm{A} = \left[ \begin{array}{cccc} \mathbb{1} & -\prescript{}{W}{\mathbf{v}_1} & \mathbb{0}&  \mathbb{0}\\
    &\vdots&\\
    \mathbb{1} &  \mathbb{0} & -\prescript{}{W}{\mathbf{v}_i} &  \mathbb{0}\\
    &\vdots&\\
    \mathbb{1} & \mathbb{0}& \mathbb{0} &  -\prescript{}{W}{\mathbf{v}_P} \end{array} \right], \,
    b = \left[ \begin{array}{c} \prescript{}{W}{\mathbf{p}_{C_1}} \\ \vdots \\  \prescript{}{W}{\mathbf{p}_{C_i}} \\ \vdots \\ \prescript{}{W}{\mathbf{p}_{C_P}}\end{array} \right],
\end{equation}
where $\prescript{}{W}{\mathbf{v}_i}$ is the $i$th centroid's bearing vector, $\mathrm{R}^{C_i}_W$ is the rotation part of $\mathrm{T}^{C_i}_W$, $Bp$ is the back-projection function in Equation~(\ref{eq:backprojection}), $\mathbb{1}$ is the identity matrix, and $P$ is the number of observations and corresponding poses.
The triangulation point $\prescript{}{W}{\mathbf{\hat{p}}}_{SQ}$ is then extracted from the first three elements of $\mathbf{x}$ obtained by solving $\mathrm{A} \cdot \mathbf{x} = \mathbf{b}$ using QR-decomposition.

Furthermore, a rough prior for the depth parameter $d_i \in \mathcal{D}$ utilized in the back-projection in Equation~(\ref{eq:backprojection}) for each camera view, i.e. a unified depth for a mask observation, is obtained by solving
\begin{equation}
    \prescript{}{W}{\mathbf{p}}_{C_i} + d_i \cdot \prescript{}{W}{\mathbf{v}_i} = \prescript{}{W}{\mathbf{\hat{p}}}_{SQ}.
\end{equation}
\subsubsection{Orientation, size, and separate depth}
To get a prior and initialization on the remaining parameters of a quadric and good estimates for a per-sample depth, we utilize a \ac{pca}-like initialization procedure. 

The main idea is to find the most prominent directions and corresponding sizes in a \ac{pc} built by back-projecting the mask observation samples $\mathcal{S}$ with the depth prior $\mathcal{D}$ estimated in the previous triangulation step. 
This is achieved by determining the principle components of the \ac{pc} using \ac{pca}~\cite{Shlens2014AAnalysis} providing a rotation matrix corresponding to the most prominent direction, i.e. the orientation of the \ac{pc}, which is used as prior for the orientation of the \ac{sq}.
Evaluating the boundaries of the rotated \ac{pc} directly serves as a prior for the object size.

Finally, a per-sample depth is obtained by optimizing $G_4$ with respect to only the depth parameter using the initialized fixed \ac{sq}, in this case a quadric with $\varepsilon = \mathbf{1}$, position determined from triangulation, and orientation and size estimated using the procedure described above.

The overall algorithm to initialize orientation, size, and separate depth is summarized in Algorithm~\ref{alg:pca}.

\begin{algorithm}[t]
\begin{algorithmic}[1]
\Procedure{PCAI}{$\mathcal{S}, \mathcal{D}, \mathcal{P}, \prescript{}{W}{\mathbf{p}_{SQ}}$}
\Algphase{Phase 1 - Back-project semantic measurement samples}
\State $\prescript{}{W}{\mathcal{T}} \gets$ \O \Comment{Initialize points in world coordinates}
\ForAll{$o \in \mathcal{O}$}
    \State $d \gets \mathcal{D}_o$ \Comment{Same depth for every sample}
    \State $\mathrm{T}_{W}^{C} \gets \mathcal{P}_o$
    \ForAll{$\mathbf{s} \in \mathcal{S}_o$}
        \State Add $\mathrm{T}_{W}^{C} \cdot Bp(\mathbf{s}, d)$ to $\prescript{}{W}{\mathcal{T}}  $ \Comment{Using Eq.~(\ref{eq:backprojection})}
    \EndFor
\EndFor
\State $N \gets \text{number of samples in } \prescript{}{W}{\mathcal{T}}$
\Algphase{Phase 2 - PCA}
\State $\mathrm{A} \gets \text{zeroMean}(\prescript{}{W}{\mathcal{T}})$
\State $\mathrm{U} \cdot \mathrm{\Sigma} \cdot \mathrm{V}^* = \svd(\frac{\mathrm{A}}{\sqrt{N-1}})$
\State $\mathrm{S} \gets \mathrm{V}^{*\top}\cdot \mathrm{A}$ \Comment{Rotated samples}
\Algphase{Phase 3 - Extract size, orientation, and depths}
\State $\mathbf{a} \gets \frac{\max{\mathrm{S}} - \min{\mathrm{S}}}{2}$
\State $\prescript{}{W}{\mathbf{r}_{SQ}} \gets \Euler(\mathrm{V^*})$ \Comment{Convert $\mathbf{\mathrm{V}^*}$ to Euler angles}
\State $\mathbf{\xi} = \left[ \mathbf{a}, \mathbf{1}, \prescript{}{W}{\mathbf{p}_{SQ}}, \prescript{}{W}{\mathbf{r}_{SQ}} \right]$ \Comment{Quadric with $\varepsilon = \mathbf{1}$}
\ForAll{$o \in \mathcal{O}$}
    \ForAll{$\mathbf{s}_i \in \mathcal{S}_o$}
        \State $d_i \in \mathcal{D} \gets \argmin_d{G_4 \left(\mathbf{\xi},  \mathbf{s}_i, \mathcal{P}_o,d\right)^2} $
    \EndFor
\EndFor
\State \Return $\left[\mathbf{\xi}, \mathcal{D}\right]$
\EndProcedure
\end{algorithmic}
\caption{PCA initialization}
\label{alg:pca}
%\vspace{-0.6cm}
\end{algorithm}

\subsubsection{Optimization}
The final step of \ac{sq} retrieval is to use the optimization introduced in Equation~(\ref{eq:opt_riou}) or~(\ref{eq:opt_d}) initialized using the previous stages.
Multiple optimization setups are used, concatenated, and compared:
\begin{enumerate}[label=\Alph*)]
    \item % optimizeRIOUSQ
    Numeric optimization of all \ac{sq} parameters using $G_2$ in Equation~(\ref{eq:opt_riou}).
    \item % optimizeRIOUQ
    Numeric optimization with $G_2$ of only quadric parameters, fixing $\varepsilon = \mathbf{1}$.
    \item % optimizeCDQ
    Optimization of quadric parameters, fixing $\varepsilon = \mathbf{1}$ and using combined depths, i.e. only optimizing one depth parameter per view, using $G_5$ in Equation~(\ref{eq:opt_d}).
    \item % optimizeSDQ
    Optimizing only quadric parameters as in option C), but optimizing a separate depth variable per sample.
    \item % optimizeSDSQ
    Optimizing all \ac{sq} parameters with separate depth variables using $G_5$ in Equation~(\ref{eq:opt_d}). This corresponds to the same potential outcome as in option A), but uses the analytic cost function $G_5$ instead of $G_2$.
    \item % optimizeRIOUs
    Using the numeric cost function $G_2$ for optimizing the shape parameters $\varepsilon$ while fixing all other parameters.
\end{enumerate}

\section{Experiments}
To evaluate the proposed method's applicability, performance, and accuracy, we conducted different preliminary experiments in a simulation environment.
\subsection{Simulation and experimental setup}
Our simulator implemented in MATLAB is able to create random \ac{gt} \ac{sq} objects, create pinhole camera trajectories that observe the object, and calculate the semantic mask observations by projecting the \ac{sq} to the camera view.
The projected \ac{sq} measurements are then randomly sampled, as described in Section~\ref{sec:cost_anaytic}, to obtain the set of semantic mask observations $\mathcal{O}$.
Like this, we can create a realistic scenario of a robotic agent or person equipped with a camera moving around an area while observing an object.
A main benefit of using the simulation environment is the decoupling of our evaluation from uncertainties and challenges typically introduced with other computer vision tasks, such as pose estimation, semantic object detection and instance segmentation, and multi-view data association, which allows for an independent investigation of the proposed method.

For the preliminary results in Section~\ref{sec:results}, we utilized three camera views arranged in a circle around the object (see Figure~\ref{fig:multi-cam-ex1}).
To evaluate the ability and robustness of fitting a \ac{sq} to multi-view mask observations, a permutation experiment is conducted by generating \ac{gt} \acp{sq} with random parameters within a given working area to ensure valid observations from the camera views.
The most important parameters of the simulation environment and permutation experiment are summarized in Table~\ref{tab:simulation}.
Finally, the camera poses and mask observations are utilized in the fitting procedure introduced in Section~\ref{sec:method} with different settings and concatenations to retrieve \acp{sq} that can be compared to the \ac{gt}.
The numeric cost functions in \texttt{Stages 3A, 3B}, and \texttt{3F} utilize the convex hull of the semantic measurement samples $\mathcal{O}$ as an observation.
For the non-linear optimization detailed in \texttt{Stage 3}, we used the Levenberg-Marquart implementation provided by the \texttt{lsqcurvefit} function.

The evaluation metrics are (1) the \ac{iou} of the estimated 3D \ac{sq} and the \ac{gt} \ac{sq}, (2) the mean 2D \ac{riou} of the re-projected estimated \ac{sq} and re-projected \ac{gt} \ac{sq} on all utilized observer camera views (R-IOU), and (3) the mean 2D \ac{riou} of the estimated \ac{sq} and the convex hull of the semantic measurements samples $\mathcal{O}$ as obtained with the procedure described in Section~\ref{sec:cost_anaytic}, termed R-IOU-M.
In addition, we also define a success rate $\sigma$.
A fit is deemed successful if the estimated and \ac{gt} \ac{sq} are overlapping, i.e. $\text{IOU} > 0$.

As the optimization is based on mask observations, the \ac{iou} might be misleading as it depends on the observability of the different parameters.
In contrast, the \ac{riou} more accurately represents the fit quality of the observation data.

\begin{table}[]
    \centering
    \caption{Overview of the simulation environment parameters}
    \begin{tabular}{lc}
    \toprule
    Camera resolution & $\unit[640 \times 480]{px}$  \\
    Camera focal length & $\unit[400 \times 300]{px}$\\
    Camera image center & $\unit[320 \times 240]{px}$\\
    Camera trajectory circle radius &   $\unit[10]{m}$\\
    Camera views    &   $3$\\
    \midrule
    Number of permutations & $100$ \\ 
    Samples per observation $N$& $100$ \\
    Constraint violation penalty $G_2$ & $c_p = 1$ \\
    Constraint violation penalty $G_3$ to $G_5$ & $c_p = 100$ \\
    \midrule
    \ac{sq} size range & $\mathbf{a} \in \unit[\{0.1,5\}]{m}$ \\
    \ac{sq} position range & $\prescript{}{W}{\mathbf{p}} \in \unit[\{-5,5\}]{m}$ \\
    \ac{sq} orientation range & $\prescript{}{W}{\mathbf{r}} \in \unit[\{-\pi, \pi\}]{rad}$ \\
    \ac{sq} shape range & $\mathbf{\varepsilon} \in \{0.1, 1.9\}$ \\
    \bottomrule
    \end{tabular}
    \label{tab:simulation}
\end{table}

\subsection{Results and Discussion} \label{sec:results}
\begin{table*}[]
    \caption{Overview of the achievable accuracy and robustness in the permutation experiment using different stage concatenations fitting 100 random \acp{sq} using multi-view mask observations. The values correspond to the final result after the last stage. M: Median, A: Average, StD: Standard deviation}
    \label{tab:results-stages}
    \centering
    \begin{tabular}{l|ccc|ccc|ccc|ccc|c}
    \toprule
    &\multicolumn{3}{c|}{IOU}&\multicolumn{3}{c|}{R-IOU}&\multicolumn{3}{c|}{R-IOU-M}&\multicolumn{3}{c|}{Time$^\text{a}$ $[\unit{s}]$}&$\sigma$\\
    \texttt{Stages} &M&A&Std&M&A&Std&M&A&Std&M&A&Std \\
    \midrule
    \texttt{3A}&0.401&0.370&0.351&0.726&0.489&0.385&0.770&0.523&0.417&2.669&6.351&8.990&$\unit[66]{\%}$\\
    \texttt{3B}&0.282&0.363&0.336&0.745&0.509&0.371&0.773&0.537&0.396&1.963&3.138&3.807&$\unit[67]{\%}$\\
    \texttt{1}\arrow\texttt{3A}&0.686&0.636&0.185&0.846&0.808&0.104&0.920&0.872&0.117&6.607&9.518&8.446&$\unit[100]{\%}$\\
    \texttt{1}\arrow\texttt{2}\arrow\texttt{3A}&0.741&0.688&0.174&0.852&0.841&0.073&0.919&0.906&0.082&5.183&7.231&5.926&$\unit[100]{\%}$\\
    \texttt{1}\arrow\texttt{2}\arrow\texttt{3B}&0.739&0.694&0.146&0.835&0.836&0.025&0.897&0.895&0.029&3.184&4.630&3.682&$\unit[100]{\%}$\\
    \midrule
    \texttt{3E}&0.000&0.000&0.001&0.007&0.015&0.021&0.007&0.015&0.021&1.075&1.066&0.277&$\unit[18]{\%}$\\
    \texttt{1}\arrow\texttt{3E}&0.009&0.044&0.071&0.089&0.137&0.142&0.103&0.154&0.156&1.009&1.018&0.263&$\unit[100]{\%}$\\
    \texttt{1}\arrow\texttt{2}\arrow\texttt{3C}&0.012&0.061&0.103&0.107&0.153&0.141&0.123&0.175&0.158&0.514&1.045&1.267&$\unit[100]{\%}$\\
    \texttt{1}\arrow\texttt{2}\arrow\texttt{3D}&0.516&0.488&0.261&0.766&0.730&0.159&0.668&0.650&0.134&0.876&1.396&3.097&$\unit[100]{\%}$\\
    \texttt{1}\arrow\texttt{2}\arrow\texttt{3E}&0.435&0.427&0.198&0.641&0.635&0.143&0.649&0.624&0.130&1.350&1.855&5.862&$\unit[100]{\%}$\\
    \midrule
    \texttt{1}\arrow\texttt{2}\arrow\texttt{3D}\arrow\texttt{3E}&0.516&0.488&0.261&0.766&0.730&0.159&0.668&0.650&0.134&0.877&1.426&3.131&$\unit[100]{\%}$\\
    \texttt{1}\arrow\texttt{2}\arrow\texttt{3D}\arrow\texttt{3A}&0.760&0.694&0.169&0.856&0.847&0.059&0.924&0.910&0.069&5.688&7.765&6.050&$\unit[100]{\%}$\\
    \texttt{1}\arrow\texttt{2}\arrow\texttt{3D}\arrow\texttt{3F}&0.622&0.536&0.234&0.784&0.749&0.126&0.793&0.759&0.127&1.895&2.837&5.706&$\unit[100]{\%}$\\
    \bottomrule
    %% only eps < 0.15 or eps > 1.85
    %Stage123A&0.73172&0.64543&0.25458&0.85092&0.85415&0.02760&0.92951&0.91720&0.02031\\
    %Stage123B&0.69458&0.60382&0.23792&0.81896&0.82507&0.02700&0.87533&0.87750&0.02007\\
    \end{tabular}\\ \vspace{0.1cm}
    \footnotesize{$^\text{a}$ The system is tested on an AMD Ryzen 5 1600 $\unit[3.2]{GHz}$, 12 threads, $\unit[16]{GB}$ RAM running MATLAB R2020a.}
\end{table*}%
The optimization of \ac{sq} parameters to fit multi-view mask observations by directly maximizing the \ac{riou}, as described in Section~\ref{sec:riou} and \texttt{Stages 3A} and \texttt{3B}, is challenging as there are zero gradients if there is no overlap between the initialized \ac{sq} and the mask observation.
This results in a low success rate and demands for a good initialization.

This is even more drastic when using the analytic cost functions described in \texttt{Stages 3C} to \texttt{3E} for the direct optimization of \ac{sq} parameters, which highly depends on a good initialization and often results in local minima and a bad fit.

We, therefore, propose different concatenations of the stages introduced in Section~\ref{sec:multi-stage} to achieve convergence.
Table~\ref{tab:results-stages} shows an overview of the different fitting setups tested.

Using the stage concatenation \texttt{1}\arrow\texttt{2}\arrow\texttt{3D}\arrow\texttt{3A} achieves the best quality as it optimally utilizes all available information, i.e. the semantic mask measurements, and is well initialized.
However, the optimization is relatively slow as it depends on finite differences for the optimization.
Note, that full parallelization is utilized for finite differences while the analytic Jacobians are evaluated on a single CPU thread.

Optimizing all \ac{sq} parameters instead of only quadrics only gives a slight advantage.
However, this evaluation is biased as the permutation uniformly samples all possible $\varepsilon$, leading to non-extreme \acp{sq} in many cases.
When specifically considering more extreme \acp{sq}, e.g. only if one of the $\varepsilon < 0.15$ or $\varepsilon > 1.85$, the performance improvement becomes more evident, i.e. 0.695 vs. 0.732, 0.819 vs. 0.851, and 0.875 vs. 0.930 for the median \ac{iou}, \ac{riou} and R-IOU-M for the combinations \texttt{1}\arrow\texttt{2}\arrow\texttt{3B} and \texttt{1}\arrow\texttt{2}\arrow\texttt{3A}, respectively.

Using the analytic cost function $G_5$ in \texttt{1}\arrow\texttt{2}\arrow \texttt{3D} shows a good performance in retrieving quadric parameters with a lower computational cost.
However, adding the shape parameters in the optimization in \texttt{Stage 3E} does not significantly improve the results.
It may even lead to the creation and convergence into new local minima.

A compromise between shape recovery and higher computational performance is achieved with the stage combination \texttt{1}\arrow\texttt{2}\arrow \texttt{3D}\arrow\texttt{3F}.
Furthermore, initializing \texttt{Stage 3A} with \texttt{3D} in the sequence \texttt{1}\arrow\texttt{2}\arrow \texttt{3D}\arrow\texttt{3A} leads to a slight boost in computational performance for the final step, but considering the time required for \texttt{Stage 3D}, which runs on a single thread, a more in-depth analysis is required to justify the computational advantage.

Finally, utilizing a single depth per camera view as proposed in \texttt{Stage 3C} did not work well as the optimization typically converges to a thin camera-aligned quadric.
This is especially evident in larger \acp{sq}.
The main reason for this is the limited flexibility in depth estimation. 

\begin{figure}
    \centering
    \includegraphics[width=1\columnwidth]{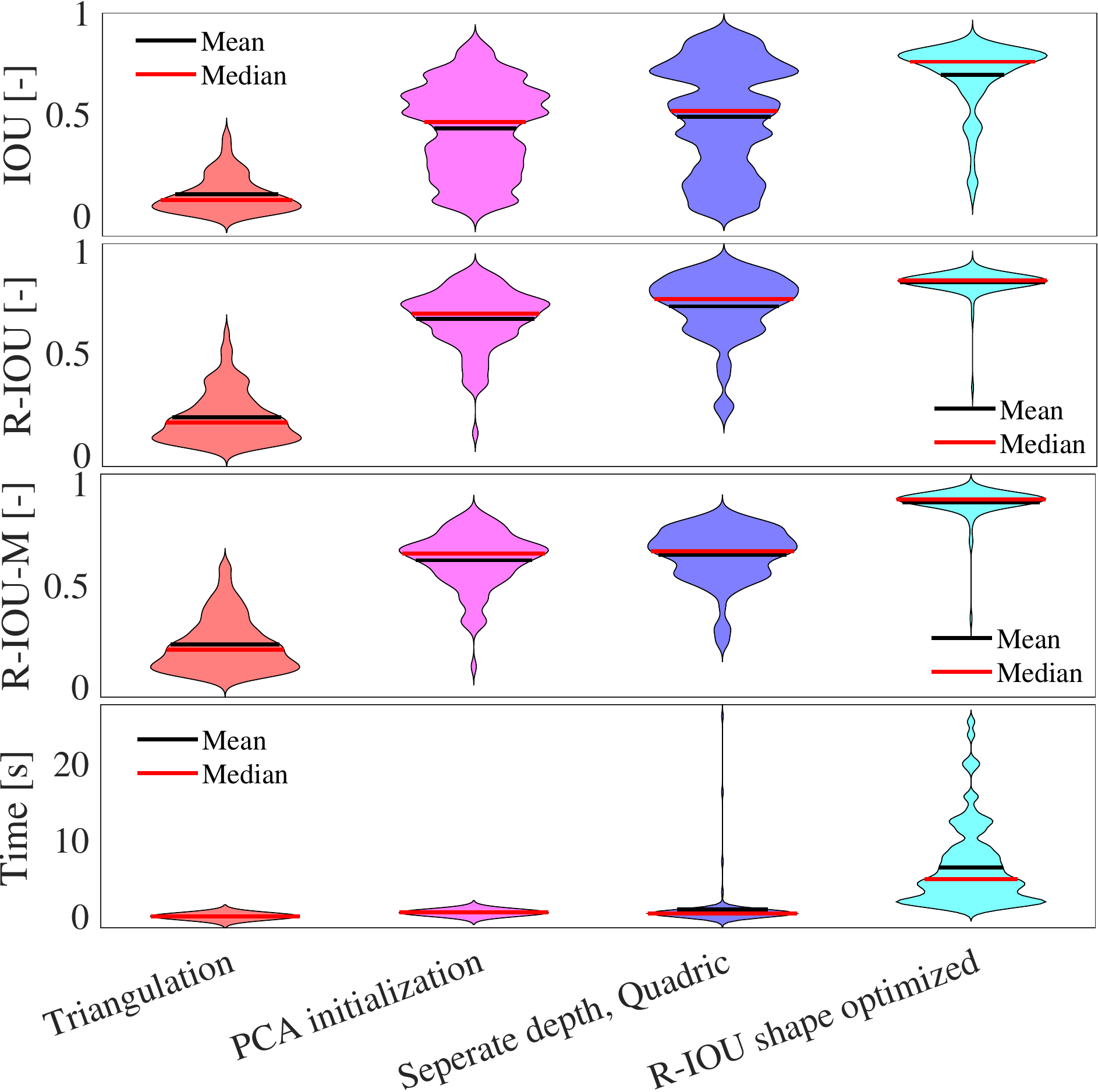}
    \caption{Result distribution of the evaluation metrics for 100 random \acp{sq} after different stages for the stage combination \texttt{1}\arrow\texttt{2}\arrow\texttt{3D}\arrow\texttt{3A}. The wider the violins, the more samples fall in that region.}
    \label{fig:result-multi-stage}
\end{figure}

\begin{figure}[bt!]
    \newcolumntype{C}[1]{>{\centering\let\newline\\\arraybackslash\hspace{0pt}}m{#1}}
    \def\colsize{1\columnwidth}
    \centering
    \begin{tabular}{C{\colsize}}
        \includegraphics[clip, width=\colsize,trim=3cm 1.cm 4cm 5cm]{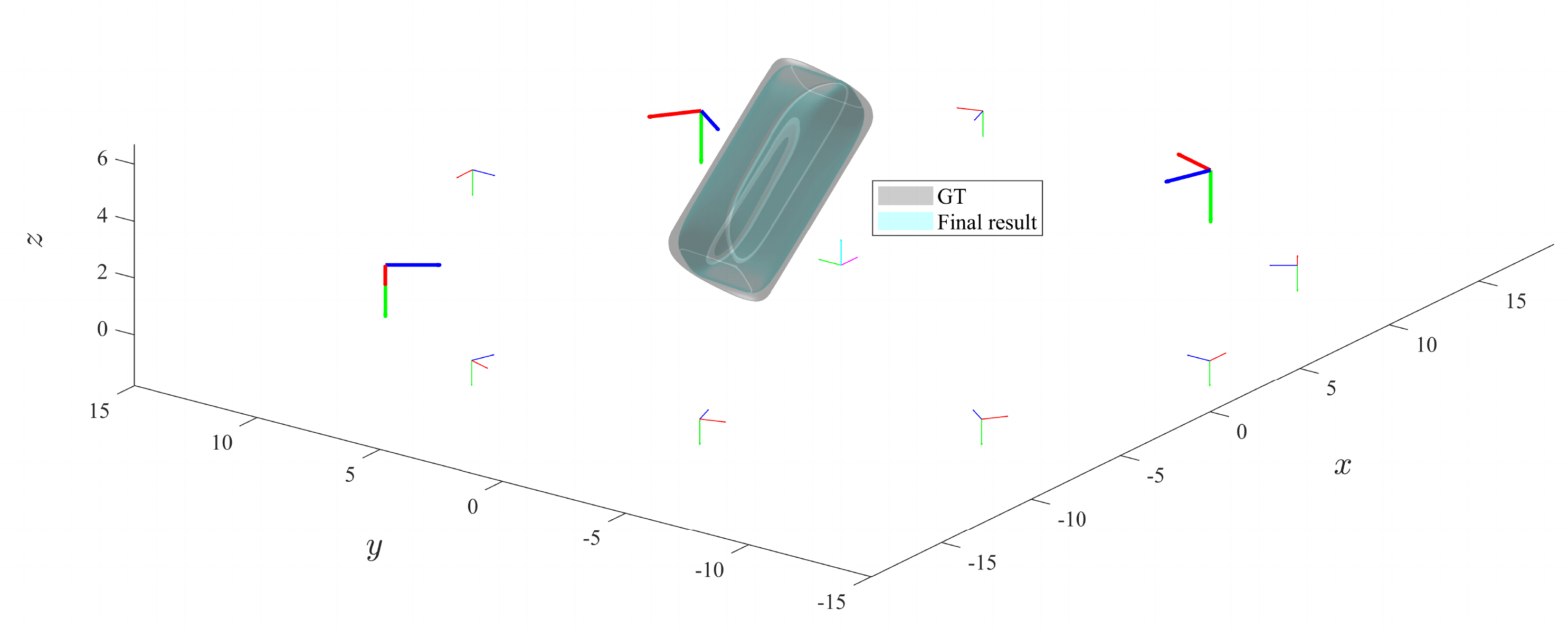}
        \\
        \footnotesize{(\textbf{a}) Overview of the final result. The cameras whose observations are utilized in the optimization are highlighted with bold coordinate frames.}\\
        \includegraphics[clip, width=\colsize,trim=1cm 0cm 1cm 0cm]{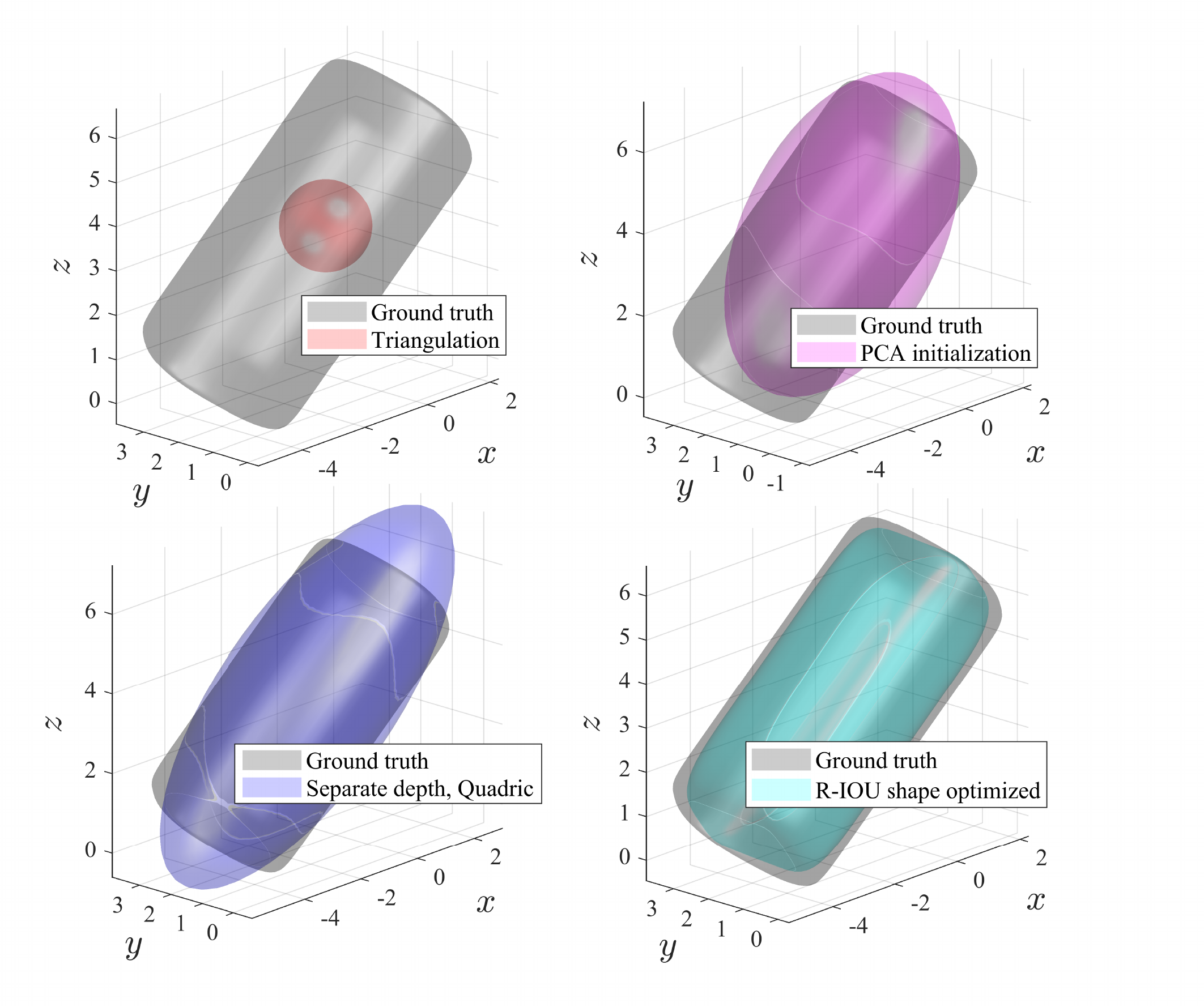}\\
        \footnotesize{(\textbf{b}) Estimation and \ac{gt} after different stages of the optimization.}
    \end{tabular}
    \caption{Example \ac{sq} retrieval for the stage combination \texttt{1}\arrow\texttt{2}\arrow\texttt{3D}\arrow\texttt{3A}.}
    \label{fig:multi-cam-ex1}
\end{figure}

Figure~\ref{fig:result-multi-stage} shows the permutation summary, and Figure~\ref{fig:multi-cam-ex1} shows an example of a typical random \ac{sq} fitted from multi-view mask observations using the stage combination \texttt{1}\arrow\texttt{2}\arrow\texttt{3D}\arrow\texttt{3A}.
Both figures showcase the individual contribution of each stage to the final solution.
\texttt{Stage 1} of the initialization, i.e triangulation, correctly recovers the average position and predicts the approximate depth parameters for each camera view. 
However, as expected, the shape, size, and orientation are far different from the \ac{gt}.
After the second stage of the optimization, the \ac{pca} initialization, orientation and size are initialized, and the depth samples are fitted to the surface of the prior quadric.
The parameters of the quadric can further be refined by performing separate depth parameter optimization in \texttt{Stage 3D}.  
The size, orientation, and position of the object are correctly recovered from only three camera observations.
Finally, \texttt{Stage 3A} achieves an accurate fit to the \ac{gt} \ac{sq} with a large \ac{iou} and \ac{riou}.

Typically, the optimized \ac{sq} size and edginess parameters are slightly underestimated, mainly due to the chosen random sampling procedure of the semantic mask observations.
This results in under-represented extremes of the contour of the re-projected \ac{sq} and is one of the reasons why \ac{sq} are only performing marginally better than quadrics in our experiments.
An adaptive sampling procedure focusing on the most informative samples might help mitigate the underestimation and make the difference between quadric and \ac{sq} parameter fitting clearer.

\section{Conclusions and Outlook}

In this work, we propose to use \acp{sq} to represent semantic objects for use in object-based semantic \ac{slam}.
We present numeric and analytic cost functions and introduce a multi-stage fitting procedure.
With our novel approach, \acp{sq} are retrieved from multi-view semantic mask observations from a monocular camera exploring an environment.
We show in preliminary experiments using various optimization setups in a simulation environment that \ac{sq} parameters are successfully recovered achieving high \acp{iou} and \acp{riou}.

However, the performance is not optimal, and the benefit \acp{sq} provide compared to standard quadrics is not always significant.
Nevertheless, we see a lot of potential for robustness and accuracy improvement.
The investigation of other non-linear optimization approaches, such as DIRECT~\cite{Jones1993LipschitzianConstant} or StoGO~\cite{Madsen1998GlobalBranch-and-Bound}, might improve the fragility with respect to local minima.
However, as most optimization-based \ac{slam} frameworks are based on Levenberg-Marquart optimization, the inclusion in such frameworks might become more challenging.
Furthermore, the optimal cost function might not have been found yet and requires further investigation.
Additional combinations and adaptions of the proposed cost functions and sampling procedures might lead to a faster convergence, increased robustness, and higher accuracy.
Finally, the extension to real-world data and full optimization of \ac{sq} parameters together with camera poses, which would allow the inclusion in a \ac{slam} framework, are subject to our current and future work. 

Nevertheless, we believe that \acp{sq} are a promising representation of semantic objects in an environment enabling appearance-invariant loop-closures and localizations~\cite{Ok2019RobustNavigation,Frey2019EfficientSLAM} due to its compact and versatile parametrization.
A semantically enriched map is easier to interpret for a human operator compared to traditional maps used in \ac{slam} and provides improved situational awareness for high-level tasks such as motion planning or manipulation.

\begin{acronym}
\acro{adas}[ADAS]{advanced driving assistance systems}
\acro{ae}[AE]{auto-exposure}
\acro{asl}[ASL]{Autonomous Systems Lab}
\acro{ba}[BA]{bundle adjustment}
\acro{bm}[BM]{block-matching}
\acro{bow}[BoW]{Bag-of-Words}
\acro{brisk}[BRISK]{Binary Rotation Invariant Scalable Keypoint}
\acro{clahe}[CLAHE]{contrast limiting adaptive histogram equalization}
\acro{cnn}[CNN]{Convolutional Neural Network}
\acro{cpu}[CPU]{central processing unit}
\acro{davis}[DAVIS]{Dynamic and Active Vision Sensor}
\acro{dcnn}[DCNN]{Deep Convolutional Neural Network}
\acro{dl}[DL]{deep learning}
\acro{dof}[DoF]{degrees of freedom}
\acro{dso}[DSO]{Direct Sparse Odometry}
\acro{dvs}[DVS]{Dynamic Vision Sensor}
\acro{ekf}[EKF]{extended Kalman filter}
\acro{etcs}[ETCS]{European Train Control System}
\acro{etsc}[ETSC]{European Train Security Council}
\acro{esdf}[ESDF]{Euclidean signed distance field}
\acro{fast}[FAST]{Features form Accelerated Segment Test}
\acro{fc}[FC]{fully connected}
\acro{fir}[FIR]{finite impulse response}
\acro{fov}[FoV]{field of view}
\acro{fpga}[FPGA]{field-programmable gate array}
\acro{fps}[FPS]{frames per second}
\acro{gnss}[GNSS]{global navigation satellite system}
\acro{gp}[GP]{Gaussian Process}
\acro{gpm}[GPM]{Gaussian preintegrated measurement}
\acro{gps}[GPS]{Global Positioning System}
\acro{gpu}[GPU]{graphics processing unit}
\acro{gt}[GT]{ground truth}
\acro{gtc}[GTC]{ground truth clustering}
\acro{gtsam}[GTSAM]{Georgia Tech Smoothing and Mapping library}
\acro{hdr}[HDR]{High Dynamic Range}
\acro{hs}[HS]{Hough space}
\acro{ht}[HT]{Hough transform}
\acro{i2c}[I$^2$C]{Inter-Integrated Circuit}
\acro{idol}[IDOL]{IMU-DVS  Odometry  with Lines}
\acro{imu}[IMU]{inertial measurement unit}
\acro{ins}[INS]{inertial navigation system}
\acro{iou}[IOU]{intersection over union}
\acro{kf}[KF]{Kalman filter}
\acro{led}[LED]{light emitting diode}
\acro{lidar}[LiDAR]{Light Detection and Ranging sensor}
\acro{lsd}[LSD]{line segment detector}
\acro{lssvm}[LSSVM]{least squares support vector machine}
\acro{lwir}[LWIR]{long-wave infrared}
\acro{mav}[MAV]{micro aerial vehicle}
\acro{mcu}[MCU]{micro controller unit}
\acro{nclt}[NCLT]{North Campus Long-Term}
\acro{nmi}[NMI]{normalized mutual information}
\acro{nms}[NMS]{non-maxima suppression}
\acro{nn}[NN]{nearest neighbor}
\acro{os}[OS]{operating system}
\acro{pcb}[PCB]{printed circuit board}
\acro{pc}[PC]{point cloud}
\acro{pca}[PCA]{principal component analysis}
\acro{pcm}[PCM]{probabilistic curvemap}
\acro{pf}[PF]{particle filter}
\acro{pps}[PPS]{Pulse per second}
\acro{ptp}[PTP]{Precision Time Protocol}
\acro{ransac}[RANSAC]{random sample consensus}
\acro{rgbdi}[RGB-D-I]{Color-Depth-Inertial}
\acro{riou}[R-IOU]{reprojection intersection over union}
\acro{rmse}[RMSE]{root mean square error}
\acro{rnn}[RNN]{recurrent neural network}
\acro{roi}[ROI]{region of interest}
\acro{ros}[ROS]{Robot Operating System}
\acro{rqe}[RQE]{Rényi's Quadric Entropy}
\acro{rtk}[RTK]{real time kinematics}
\acro{sbb}[SBB]{Schweizerische Bundesbahnen}
\acro{sift}[SIFT]{Scale Invariant Feature Transform}
\acro{slam}[SLAM]{Simultaneous Localization And Mapping}
\acro{snn}[SNN]{spiking neural network}
\acro{snr}[SNR]{signal to noise ratio}
\acro{sp}[SP]{SuperPoint}
\acro{spi}[SPI]{Serial Peripheral Interface}
\acro{sq}[SQ]{superquadric}
\acro{swe}[SWE]{Sliding Window Estimator}
\acro{tof}[ToF]{time of flight}
\acro{tsdf}[TSDF]{Truncated Signed Distance Function}
\acro{tum}[TUM]{Technische Universit\"at M\"unchen}
\acro{uart}[UART]{Universal Asynchronous Receiver Transmitter}
\acro{uav}[UAV]{unmanned aerial vehicle}
\acro{usb}[USB]{universal serial bus}
\acro{vbg}[VBG]{Verkehrsbetriebe Glattal AG}
\acro{vbz}[VBZ]{Verkehrsbetriebe Zürich}
\acro{VersaVIS}[VersaVIS]{Open Versatile Multi-Camera Visual-Inertial Sensor Suite}
\acro{vi}[VI]{visual-inertial}
\acro{vio}[VIO]{visual-inertial odometry}
\acro{vo}[VO]{visual odometry}
\acro{zvv}[ZVV]{Z\"urich Verkehrsverein}
\acro{svd}[SVD]{Singular Value Decomposition}
\acro{dlt}[DLT]{Direct Linear Transform}
\acro{ugv}[UGV]{unmanned ground vehicle}
\acro{ar}[AR]{augmented reality}
\acro{vr}[VR]{virtual reality}
\acro{ato}[ATO]{autonomous train operation}
\end{acronym}

%\section*{References}
\bibliographystyle{IEEEtran}
\bibliography{references}

\begin{thebibliography}{10}
\providecommand{\url}[1]{#1}
\csname url@samestyle\endcsname
\providecommand{\newblock}{\relax}
\providecommand{\bibinfo}[2]{#2}
\providecommand{\BIBentrySTDinterwordspacing}{\spaceskip=0pt\relax}
\providecommand{\BIBentryALTinterwordstretchfactor}{4}
\providecommand{\BIBentryALTinterwordspacing}{\spaceskip=\fontdimen2\font plus
\BIBentryALTinterwordstretchfactor\fontdimen3\font minus
  \fontdimen4\font\relax}
\providecommand{\BIBforeignlanguage}[2]{{%
\expandafter\ifx\csname l@#1\endcsname\relax
\typeout{** WARNING: IEEEtran.bst: No hyphenation pattern has been}%
\typeout{** loaded for the language `#1'. Using the pattern for}%
\typeout{** the default language instead.}%
\else
\language=\csname l@#1\endcsname
\fi
#2}}
\providecommand{\BIBdecl}{\relax}
\BIBdecl

\bibitem{Gawel2019AConstruction}
A.~Gawel, R.~Siegwart, M.~Hutter, T.~Sandy, H.~Blum, J.~Pankert, K.~Kramer,
  L.~Bartolomei, S.~Ercan, F.~Farshidian, M.~Chli, and F.~Gramazio, ``{A
  Fully-Integrated Sensing and Control System for High-Accuracy Mobile Robotic
  Building Construction},'' in \emph{IEEE International Conference on
  Intelligent Robots and Systems}.\hskip 1em plus 0.5em minus 0.4em\relax
  Institute of Electrical and Electronics Engineers Inc., 11 2019, pp.
  2300--2307.

\bibitem{Blosch2010VisionEnvironments}
M.~Bl{\"{o}}sch, S.~Weiss, D.~Scaramuzza, and R.~Siegwart, ``{Vision based MAV
  navigation in unknown and unstructured environments},'' in \emph{Proceedings
  - IEEE International Conference on Robotics and Automation}, 2010, pp.
  21--28.

\bibitem{Burki2019}
M.~B{\"{u}}rki, C.~Cadena, I.~Gilitschenski, R.~Siegwart, and J.~Nieto,
  ``{Appearance‐based landmark selection for visual localization},''
  \emph{Journal of Field Robotics}, vol.~36, no.~6, pp. 1041--1073, 9 2019.

\bibitem{Tschopp2019ExperimentalVehicles}
F.~Tschopp, T.~Schneider, A.~W. Palmer, N.~Nourani-Vatani, C.~Cadena,
  R.~Siegwart, and J.~Nieto, ``{Experimental comparison of visual-aided
  odometry methods for rail vehicles},'' \emph{IEEE Robotics and Automation
  Letters}, vol.~4, no.~2, pp. 1815--1822, 2019.

\bibitem{RolandSiegwart2011}
R.~Siegwart, I.~R. Nourbakhsh, and D.~Scaramuzza, \emph{{Introduction to
  Autonomous Mobile Robots}}, 2nd~ed.\hskip 1em plus 0.5em minus 0.4em\relax
  Camebridge: MIT Press, 2011.

\bibitem{Schneider2017}
T.~Schneider, M.~Dymczyk, M.~Fehr, K.~Egger, S.~Lynen, I.~Gilitschenski, and
  R.~Siegwart, ``{maplab: An Open Framework for Research in Visual-inertial
  Mapping and Localization},'' \emph{IEEE Robotics and Automation Letters},
  vol.~3, no.~3, pp. 1418--1425, 11 2018.

\bibitem{Mur-Artal2015}
\BIBentryALTinterwordspacing
R.~R. Mur-Artal, J.~M.~M. Montiel, and J.~D. Tardos, ``{ORB-SLAM: A Versatile
  and Accurate Monocular SLAM System},'' \emph{IEEE Transactions on Robotics},
  vol.~31, no.~5, pp. 1147--1163, 10 2015. [Online]. Available:
  \url{http://ieeexplore.ieee.org/document/7219438/}
\BIBentrySTDinterwordspacing

\bibitem{Leutenegger2011}
\BIBentryALTinterwordspacing
S.~Leutenegger, M.~M. Chli, and R.~Y. Siegwart, ``{Binary Robust Invariant
  Scalable Keypoints},'' in \emph{Proceedings of the IEEE International
  Conference on Computer Vision}, Barcelona, 2011, pp. 2548--2555. [Online].
  Available: \url{https://www.robots.ox.ac.uk/~vgg/rg/papers/brisk.pdf}
\BIBentrySTDinterwordspacing

\bibitem{Rublee}
E.~Rublee, V.~Rabaud, K.~Konolige, and G.~Bradski, ``{ORB: An efficient
  alternative to SIFT or SURF},'' in \emph{Proceedings of the IEEE
  International Conference on Computer Vision}, 2011, pp. 2564--2571.

\bibitem{Milford2012}
\BIBentryALTinterwordspacing
M.~J. Milford and G.~F. Wyeth, ``{SeqSLAM: Visual route-based navigation for
  sunny summer days and stormy winter nights},'' in \emph{2012 IEEE
  International Conference on Robotics and Automation}.\hskip 1em plus 0.5em
  minus 0.4em\relax IEEE, 5 2012, pp. 1643--1649. [Online]. Available:
  \url{http://ieeexplore.ieee.org/document/6224623/}
\BIBentrySTDinterwordspacing

\bibitem{Sattler2018BenchmarkingConditions}
T.~Sattler, W.~Maddern, C.~Toft, A.~Torii, L.~Hammarstrand, E.~Stenborg,
  D.~Safari, M.~Okutomi, M.~Pollefeys, J.~Sivic, F.~Kahl, and T.~Pajdla,
  ``{Benchmarking 6DOF Outdoor Visual Localization in Changing Conditions},''
  in \emph{Proceedings of the IEEE Computer Society Conference on Computer
  Vision and Pattern Recognition}.\hskip 1em plus 0.5em minus 0.4em\relax IEEE
  Computer Society, 12 2018, pp. 8601--8610.

\bibitem{Mur-Artal2017}
R.~Mur-Artal and J.~D. Tardos, ``{ORB-SLAM2: An Open-Source SLAM System for
  Monocular, Stereo, and RGB-D Cameras},'' \emph{IEEE Transactions on
  Robotics}, pp. 1--8, 2017.

\bibitem{Engel2018DirectOdometry}
J.~Engel, V.~Koltun, and D.~Cremers, ``{Direct Sparse Odometry},'' \emph{IEEE
  Transactions on Pattern Analysis and Machine Intelligence}, 2018.

\bibitem{He2017}
K.~He, G.~Gkioxari, P.~Dollar, and R.~Girshick, ``{Mask R-CNN},'' in
  \emph{Proceedings of the IEEE International Conference on Computer Vision},
  Singapore, 2017, pp. 2980--2988.

\bibitem{Bolya2020YOLACT++:Segmentation}
D.~Bolya, C.~Zhou, F.~Xiao, and Y.~J. Lee, ``{YOLACT++: Better Real-time
  Instance Segmentation},'' \emph{IEEE Transactions on Pattern Analysis and
  Machine Intelligence}, 2020.

\bibitem{Redmon2016YouDetection}
J.~Redmon, S.~Divvala, R.~Girshick, and A.~Farhadi, ``{You only look once:
  Unified, real-time object detection},'' in \emph{Proceedings of the IEEE
  Computer Society Conference on Computer Vision and Pattern Recognition}, vol.
  2016-Decem.\hskip 1em plus 0.5em minus 0.4em\relax IEEE Computer Society, 12
  2016, pp. 779--788.

\bibitem{Verhagen2014Scale-invariantMatching}
B.~Verhagen, R.~Timofte, and L.~Van~Gool, ``{Scale-invariant line descriptors
  for wide baseline matching},'' in \emph{2014 IEEE Winter Conference on
  Applications of Computer Vision, WACV 2014}.\hskip 1em plus 0.5em minus
  0.4em\relax IEEE Computer Society, 2014, pp. 493--500.

\bibitem{Lee2014OutdoorLines}
J.~H. Lee, S.~Lee, G.~Zhang, J.~Lim, W.~K. Chung, and I.~H. Suh, ``{Outdoor
  place recognition in urban environments using straight lines},'' in
  \emph{Proceedings - IEEE International Conference on Robotics and
  Automation}.\hskip 1em plus 0.5em minus 0.4em\relax Institute of Electrical
  and Electronics Engineers Inc., 9 2014, pp. 5550--5557.

\bibitem{Micusik2015DescriptorSegments}
B.~Micusik and H.~Wildenauer, ``{Descriptor free visual indoor localization
  with line segments},'' in \emph{Proceedings of the IEEE Computer Society
  Conference on Computer Vision and Pattern Recognition}, vol.
  07-12-June-2015.\hskip 1em plus 0.5em minus 0.4em\relax IEEE Computer
  Society, 10 2015, pp. 3165--3173.

\bibitem{Oliva2006BuildingGlob}
\BIBentryALTinterwordspacing
A.~Oliva and A.~Torralba, ``{Building the gist of a scene: the role ofal image
  features in recognition glob},'' \emph{Progress in Brain Research}, vol. 155
  B, pp. 23--36, 2006. [Online]. Available:
  \url{https://pubmed.ncbi.nlm.nih.gov/17027377/}
\BIBentrySTDinterwordspacing

\bibitem{Arandjelovic2013AllVLAD}
R.~Arandjelovic and A.~Zisserman, ``{All about VLAD},'' in \emph{Proceedings of
  the IEEE Computer Society Conference on Computer Vision and Pattern
  Recognition}, Portland, OR, USA, 2013, pp. 1578--1585.

\bibitem{Arandjelovic2018NetVLAD:Recognition}
R.~Arandjelovic, P.~Gronat, A.~Torii, T.~Pajdla, and J.~Sivic, ``{NetVLAD: CNN
  Architecture for Weakly Supervised Place Recognition},'' \emph{IEEE
  Transactions on Pattern Analysis and Machine Intelligence}, vol.~40, no.~6,
  pp. 1437--1451, 6 2018.

\bibitem{Gawel2017}
A.~Gawel, C.~Del~Don, R.~Siegwart, J.~Nieto, and C.~Cadena, ``{X-View:
  Graph-Based Semantic Multi-View Localization},'' \emph{IEEE Robotics and
  Automation Letters}, vol.~3, no.~3, pp. 1687--1694, 2017.

\bibitem{Taubner2020LCDRecognition}
F.~Taubner, F.~Tschopp, T.~Novkovic, R.~Siegwart, and F.~Furrer, ``{LCD - Line
  Clustering and Description for Place Recognition},'' in \emph{Proceedings -
  2020 International Conference on 3D Vision, 3DV 2020}.\hskip 1em plus 0.5em
  minus 0.4em\relax Institute of Electrical and Electronics Engineers Inc., 11
  2020, pp. 908--917.

\bibitem{Schonberger2017}
J.~L. Sch{\"{o}}nberger, M.~Pollefeys, A.~Geiger, and T.~Sattler, ``{Semantic
  Visual Localization},'' in \emph{2018 IEEE/CVF Conference on Computer Vision
  and Pattern Recognition}, Salt Lake City, UT, USA, 2017.

\bibitem{Cramariuc2021SemSegMapLocalization}
A.~Cramariuc, F.~Tschopp, N.~Alatur, S.~Benz, T.~Falck, M.~Br{\"{u}}hlmeier,
  B.~Hahn, J.~Nieto, and R.~Siegwart, ``{SemSegMap – 3D Segment-based
  Semantic Localization},'' \emph{submitted for publication}, 2021.

\bibitem{Nicholson2019QuadricSLAM:SLAM}
L.~Nicholson, M.~Milford, and N.~Sunderhauf, ``{QuadricSLAM: Dual quadrics from
  object detections as landmarks in object-oriented SLAM},'' \emph{IEEE
  Robotics and Automation Letters}, vol.~4, no.~1, pp. 1--8, 2019.

\bibitem{Frey2019EfficientSLAM}
K.~M. Frey, T.~J. Steiner, and J.~P. How, ``{Efficient constellation-based
  map-merging for semantic SLAM},'' in \emph{Proceedings - IEEE International
  Conference on Robotics and Automation}, vol. 2019-May, Montreal, Canada,
  2019, pp. 1302--1308.

\bibitem{Strasdat2010}
H.~Strasdat, J.~M.~M. Montiel, and A.~J. Davison, ``{Real-time monocular SLAM:
  Why filter?}'' in \emph{Proceedings - IEEE International Conference on
  Robotics and Automation}, 2010, pp. 2657--2664.

\bibitem{Tschopp2021Hough2Map}
F.~Tschopp, C.~Von~Einem, A.~Cramariuc, D.~Hug, A.~W. Palmer, R.~Siegwart,
  M.~Chli, and J.~Nieto, ``{Hough2Map-Iterative Event-Based Hough Transform for
  High-Speed Railway Mapping},'' \emph{IEEE Robotics and Automation Letters},
  vol.~6, no.~2, pp. 2745--2752, 4 2021.

\bibitem{Yang2019CubeSLAM:SLAM}
S.~Yang and S.~Scherer, ``{CubeSLAM: Monocular 3-D Object SLAM},'' \emph{IEEE
  Transactions on Robotics}, vol.~35, no.~4, pp. 925--938, 2019.

\bibitem{Papadakis2018}
J.~Papadakis, A.~Willis, and J.~Gantert, ``{RGBD-Sphere SLAM},'' in
  \emph{SoutheastCon 2018}.\hskip 1em plus 0.5em minus 0.4em\relax St.
  Petersburg, FL, USA: Institute of Electrical and Electronics Engineers Inc.,
  10 2018.

\bibitem{Ok2019RobustNavigation}
K.~Ok, K.~Liu, K.~Frey, J.~P. How, and N.~Roy, ``{Robust Object-based SLAM for
  High-speed Autonomous Navigation},'' in \emph{2019 International Conference
  on Robotics and Automation (ICRA)}, Montreal, Canada, 5 2019, pp. 669--675.

\bibitem{Mccormac2018Fusion++:Map}
J.~Mccormac, R.~Clark, M.~Bloesch, A.~J. Davison, and S.~Leutenegger,
  ``{Fusion++: Volumetric Object-Level SLAM Reconstructed Object Inventory Live
  Frame View Object-centric Map},'' in \emph{2018 International Conference on
  3D Vision (3DV)}, 2018, pp. 32--41.

\bibitem{Grinvald2019}
M.~Grinvald, F.~Furrer, T.~Novkovic, J.~J. Chung, C.~Cadena, R.~Siegwart, and
  J.~Nieto, ``{Volumetric instance-aware semantic mapping and 3D object
  discovery},'' \emph{IEEE Robotics and Automation Letters}, vol.~4, no.~3, pp.
  3037--3044, 7 2019.

\bibitem{Strecke2019EM-Fusion:Association}
\BIBentryALTinterwordspacing
M.~Strecke, J.~St{\"{u}}ckler, J.~Stueckler, and J.~St{\"{u}}ckler,
  ``{EM-Fusion: Dynamic Object-Level SLAM with Probabilistic Data
  Association},'' in \emph{Proceedings of the IEEE International Conference on
  Computer Vision}.\hskip 1em plus 0.5em minus 0.4em\relax Seoul, Korea:
  Institute of Electrical and Electronics Engineers Inc., 10 2019, pp.
  5865--5874. [Online]. Available: \url{http://arxiv.org/abs/1904.11781}
\BIBentrySTDinterwordspacing

\bibitem{Rosinol2020Kimera:Mapping}
A.~Rosinol, M.~Abate, Y.~Chang, and L.~Carlone, ``{Kimera: An Open-Source
  Library for Real-Time Metric-Semantic Localization and Mapping},'' in
  \emph{Proceedings - IEEE International Conference on Robotics and
  Automation}.\hskip 1em plus 0.5em minus 0.4em\relax Institute of Electrical
  and Electronics Engineers Inc., 5 2020, pp. 1689--1696.

\bibitem{Galvez-Lopez2016Real-timeSLAM}
D.~G{\'{a}}lvez-L{\'{o}}pez, M.~Salas, J.~D. Tard{\'{o}}s, and J.~M. Montiel,
  ``{Real-time monocular object SLAM},'' \emph{Robotics and Autonomous
  Systems}, vol.~75, pp. 435--449, 2016.

\bibitem{Salas-Moreno2013SLAM++:Objects}
R.~F. Salas-Moreno, R.~A. Newcombe, H.~Strasdat, P.~H. Kelly, and A.~J.
  Davison, ``{SLAM++: Simultaneous localisation and mapping at the level of
  objects},'' in \emph{Proceedings of the IEEE Computer Society Conference on
  Computer Vision and Pattern Recognition}.\hskip 1em plus 0.5em minus
  0.4em\relax Portland, OR, USA: IEEE, 2013, pp. 1352--1359.

\bibitem{Barr1981SuperquadricsTransformations}
A.~H. Barr, ``{Superquadrics and Angle-Preserving Transformations},''
  \emph{IEEE Computer Graphics and Applications}, vol.~1, no.~1, pp. 11--23,
  1981.

\bibitem{Boult1988}
T.~E. Boult and A.~D. Gross, ``{Recovery Of Superquadrics From 3-D
  Information},'' in \emph{Intelligent Robots and Computer Vision VI}, vol.
  0848.\hskip 1em plus 0.5em minus 0.4em\relax SPIE, 2 1988, p. 358.

\bibitem{Wu1994}
K.~Wu and M.~D. Levine, ``{Recovering Parametric Geons from Multiview Range
  Data},'' in \emph{1994 Proceedings of IEEE Conference on Computer Vision and
  Pattern Recognition}, Seattle, WA, USA, 1994.

\bibitem{Zhang2003}
Y.~Zhang, ``{Experimental comparison of superquadric fitting objective
  functions},'' \emph{Pattern Recognition Letters}, vol.~24, no.~14, pp.
  2185--2193, 2003.

\bibitem{Duncan2013}
K.~Duncan, S.~Sarkar, R.~Alqasemi, and R.~Dubey, ``{Multi-scale superquadric
  fitting for efficient shape and pose recovery of unknown objects},'' in
  \emph{Proceedings - IEEE International Conference on Robotics and
  Automation}.\hskip 1em plus 0.5em minus 0.4em\relax Karlsruhe, Germany: IEEE,
  2013, pp. 4238--4243.

\bibitem{Makhal2017}
\BIBentryALTinterwordspacing
A.~Makhal, F.~Thomas, and A.~P. Gracia, ``{Grasping Unknown Objects in Clutter
  by Superquadric Representation},'' in \emph{2018 Second IEEE International
  Conference on Robotic Computing (IRC)}, Laguna Hills, CA, USA, 10 2018.
  [Online]. Available: \url{http://arxiv.org/abs/1710.02121}
\BIBentrySTDinterwordspacing

\bibitem{Vaskevicius2019}
N.~Vaskevicius and A.~Birk, ``{Revisiting Superquadric Fitting: A Numerically
  Stable Formulation},'' \emph{IEEE Transactions on Pattern Analysis and
  Machine Intelligence}, vol.~41, no.~1, pp. 220--233, 1 2019.

\bibitem{Paschalidou2019}
\BIBentryALTinterwordspacing
D.~Paschalidou, A.~O. Ulusoy, and A.~Geiger, ``{Superquadrics Revisited:
  Learning 3D Shape Parsing beyond Cuboids},'' in \emph{Proceedings IEEE Conf.
  on Computer Vision and Pattern Recognition (CVPR)}, Long Beach, USA, 4 2019.
  [Online]. Available: \url{http://arxiv.org/abs/1904.09970}
\BIBentrySTDinterwordspacing

\bibitem{Jaklic2000}
A.~Jakli{\v{c}}, A.~Leonardis, and F.~Solina, ``{Superquadrics and Their
  Geometric Properties},'' in \emph{Segmentation and Recovery of Superquadrics,
  Computational Imaging and Vision}.\hskip 1em plus 0.5em minus 0.4em\relax
  Dordrecht: Springer, 2000, vol.~20, pp. 13--39.

\bibitem{Levenberg1944ASquares}
\BIBentryALTinterwordspacing
K.~Levenberg, ``{A method for the solution of certain non-linear problems in
  least squares},'' \emph{Quarterly of Applied Mathematics}, vol.~2, no.~2, pp.
  164--168, 7 1944. [Online]. Available:
  \url{https://www.ams.org/qam/1944-02-02/S0033-569X-1944-10666-0/}
\BIBentrySTDinterwordspacing

\bibitem{Marquardt1963AnParameters}
D.~W. Marquardt, ``{An Algorithm for Least-Squares Estimation of Nonlinear
  Parameters},'' \emph{Journal of the Society for Industrial and Applied
  Mathematics}, vol.~11, no.~2, pp. 431--441, 6 1963.

\bibitem{MathWorksSchweizPolyshapeMATLAB}
\BIBentryALTinterwordspacing
{MathWorks Schweiz}, ``{polyshape - 2-D polygons in MATLAB},'' 2017. [Online].
  Available: \url{https://ch.mathworks.com/help/matlab/ref/polyshape.html}
\BIBentrySTDinterwordspacing

\bibitem{Shlens2014AAnalysis}
J.~Shlens, ``{A Tutorial on Principal Component Analysis},'' Google Research,
  Mountain View, CA 94043, Tech. Rep., 4 2014.

\bibitem{Jones1993LipschitzianConstant}
\BIBentryALTinterwordspacing
D.~R. Jones, C.~D. Perttunen, and B.~E. Stuckman, ``{Lipschitzian optimization
  without the Lipschitz constant},'' \emph{Journal of Optimization Theory and
  Applications}, vol.~79, no.~1, pp. 157--181, 10 1993. [Online]. Available:
  \url{https://link.springer.com/article/10.1007/BF00941892}
\BIBentrySTDinterwordspacing

\bibitem{Madsen1998GlobalBranch-and-Bound}
K.~Madsen and S.~Zertchaninov, ``{Global Optimization using
  Branch-and-Bound},'' Department of Mathematical Modelling, Technical
  University of Denmark, Tech. Rep., 1998.

\end{thebibliography}

\end{document}